\documentclass[letterpaper, 10 pt, conference]{ieeeconf} 
\IEEEoverridecommandlockouts
\usepackage{graphicx}
\usepackage{amsmath}
\usepackage{amssymb}
\usepackage{xspace}
\usepackage{adjustbox}
\usepackage{booktabs}
\usepackage{multirow}
\usepackage{tabularx}
\usepackage[table]{xcolor}
\usepackage{subcaption}
\usepackage{cite}
\usepackage{mathptmx}
\usepackage{hyperref}
\hypersetup{
    pdfauthor={Anonymous RA-L Submission},
    pdftitle={MoMaStage: Skill-State Graph Guided Planning and Closed-Loop Execution for Long-Horizon Indoor Mobile Manipulation},
    hidelinks
}

\newcommand{\pname}{MoMaStage\xspace}

\title{\LARGE \bf \pname: Skill-State Graph Guided Planning and Closed-Loop Execution for Long-Horizon Indoor Mobile Manipulation}

\author{Chenxu Li$^{*}$, Zixuan Chen$^{*}$, Yetao Li, Jiapeng Xu, Hongyu Ding, Jieqi Shi, Jing Huo, Yang Gao
\thanks{*Equal Contribution}
\thanks{All authors are with Nanjing University, China. Emails: \{chenxuli, 241300010, 522025330117, hongyuding\}@smail.nju.edu.cn, \{chenzx, isjieqi, huojing, gaoy\}@nju.edu.cn}
}

\begin{document}

\maketitle
\thispagestyle{empty}
\pagestyle{empty}

\begin{abstract}
Long-horizon indoor mobile manipulation (MoMa) requires robots to execute extended navigation-manipulation sequences whose feasibility depends on state changes induced by preceding skills. Vision-language models (VLMs) can decompose instructions into plausible skill sequences, but they do not reliably track such cumulative embodiment constraints or revise a plan when execution deviates from expectation. We present \pname, a map-light framework for state-consistent planning and closed-loop execution in long-horizon indoor MoMa. \pname couples a frozen VLM with robot execution through three mechanisms: (i) a hierarchical library of grounded, executable skills and a topology-only projection of a Skill-State Graph (SSG) that constrains the VLM's planning space; (ii) an SSG verifier that propagates scene-region and gripper-occupancy state to reject infeasible plans before execution; and (iii) an event-driven monitor that triggers graph-grounded repair only when an observed outcome invalidates the remaining plan. The SSG captures the compact embodiment state needed for skill sequencing without requiring a dense scene map, while geometric and contact-level conditions remain within the underlying controllers. Experiments in physics-rich simulation and on a real mobile manipulator show that \pname improves planning validity and long-horizon execution survival over the evaluated baselines, while reducing latency  and model token consumption. Supplementary videos are available on our anonymous project website at \url{https://chenxuli-cxli.github.io/MoMaStage/}.
\end{abstract}
\section{INTRODUCTION}
\label{sec:introduction}

Long-horizon indoor mobile manipulation (MoMa) requires robots to execute extended sequences of navigation and object interaction in everyday environments such as kitchens and homes~\cite{yenamandra2023homerobot,chen2025owmm}. The feasibility of each skill depends on the state produced by preceding skills: the robot may move to another scene region, occupy one or both grippers, or otherwise enter a state in which the next nominal skill is no longer applicable. End-to-end imitation learning and vision-language-action (VLA) policies~\cite{zhao2023aloha,chi2023diffusion,ze2024dp3,brohan2022rt1,brohan2023rt2,octo2023,kim2024openvla} have advanced short-horizon control and visual grounding. At longer horizons, however, planning errors and execution failures compound across heterogeneous skills. 

\begin{figure}
\vspace{0.7em}
\centering
\includegraphics[width=\linewidth]{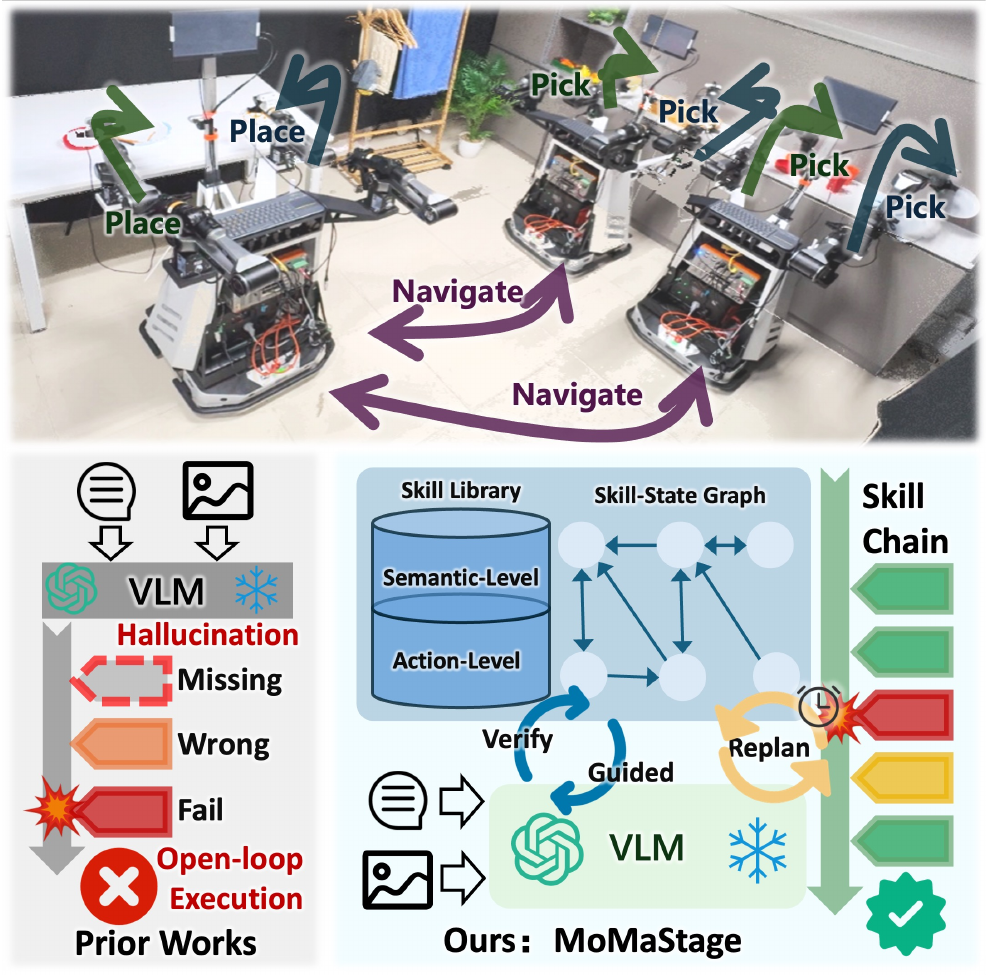}
\caption{\small \pname structures planning and execution around a frozen VLM. A hierarchical skill library and Skill-State Graph constrain and verify task decomposition, while execution feedback triggers local, graph-grounded repair when an observed outcome deviates from the expected state transition.}
\label{fig:teaser}
\vspace{-2em}
\end{figure}

Vision-language models (VLMs) provide a complementary planning layer by mapping natural-language instructions to skill sequences through commonsense and task-decomposition priors, without task-specific model training~\cite{ahn2022saycan,huang2022inner,liang2023code,liu2023llmp,singh2023progprompt}. Yet directly transferring these plans to embodied systems remains unreliable. A frozen VLM may propose unavailable skills, lose track of state changes accumulated across preceding actions, or continue from assumptions invalidated by execution failures. Querying it after every low-level action also adds unnecessary latency and token cost. These limitations create a system-interface problem in which plans must be restricted to executable skills, verified against cumulative robot state, and revised when execution feedback invalidates the remaining sequence. Scene-graph and map-based systems~\cite{rana2023sayplan,ekpo2024verigraph,uniplan2025} provide such structure through explicit environment representations, but often require geometric models or benchmark-specific infrastructure. This motivates a map-light, skill-centric interface that captures the structural dependencies needed for plan composition without representing the full environment.

We propose \textbf{\pname} (Fig.~\ref{fig:teaser}), a structured framework for state-consistent planning and closed-loop execution in long-horizon indoor MoMa. The framework assumes only a small set of named scene regions, a predefined skill library, and a manually specified skill-transition graph, without requiring a dense map or complete symbolic world model. Its design follows three principles. First, the planning interface is constrained by a hierarchical library of grounded, executable skills and a topology-only projection of a Skill-State Graph (SSG) that exposes task-relevant skill dependencies. Second, an SSG verifier propagates cumulative scene-region and gripper-occupancy state before execution and returns structured errors for plan repair. Third, replanning is event-driven rather than step-driven. Proprioceptive monitoring runs continuously, while VLM-based semantic checks are event-triggered and graph-grounded repair occurs only when an observed outcome invalidates the remaining plan. Conditions not represented in the core SSG, including receptacle state, support relations, reachability, collision avoidance, and grasp feasibility, remain the responsibility of the underlying skills and controllers. Thus, \pname targets state-consistent skill sequencing and event-triggered recovery rather than general geometric reasoning. In the evaluated settings, our experiments further suggest that explicitly exposing \textbf{skill-to-skill dependencies} is critical to reliable sequencing, whereas richer scene descriptions or additional subgoal decomposition alone do not ensure state consistency.

In summary, our primary contributions are as follows:
\begin{itemize}

\item We present \textbf{\pname}, a map-light framework that couples a frozen VLM with a hierarchical library of executable skills for closed-loop long-horizon indoor MoMa, without requiring a dense scene map or task-specific VLM training.

\item We introduce an SSG-based planning and execution mechanism that explicitly exposes task-relevant skill dependencies, verifies cumulative scene-region and gripper-occupancy state before execution, and supports targeted repair through proprioceptive monitoring and event-triggered semantic replanning.

\item We evaluate \pname in simulation and on a real dual-arm mobile manipulator using component ablations, difficulty stratification, and graph-scale analysis, and separately report planning validity, execution survival, and end-to-end task completion.

\end{itemize}
\section{RELATED WORK}
\label{sec:related}

\subsection{Mobile Manipulation and Long-Horizon Tasks}
Mobile manipulation has progressed from tabletop tasks toward household and open-vocabulary operation~\cite{li2023behavior,yenamandra2023homerobot,liu2024okrobot}. Household benchmarks such as BEHAVIOR emphasize the diversity of everyday activities and the need to evaluate behavior in virtual, interactive, and ecological settings~\cite{li2023behavior}, while HomeRobot and OK-Robot demonstrate increasingly open-vocabulary operation by combining mobile platforms with pretrained perception and language components~\cite{yenamandra2023homerobot,liu2024okrobot}. Whole-body systems~\cite{fu2024mobilealoha} extend the action space to coordinated base--arm behavior. In parallel, visuomotor policies~\cite{zhao2023aloha,chi2023diffusion} learn manipulation behaviors from demonstrations, and VLA models~\cite{brohan2022rt1,brohan2023rt2,kim2024openvla} provide increasingly capable visual grounding and action generation. These methods improve the low-level perception-control interface, but long-horizon tasks still require a mechanism to compose heterogeneous skills~\cite{chen2026agile,chenadacleargrasp,chen2026deco}, track state changes induced by earlier actions, and respond when execution deviates from the nominal sequence. The success of an individual policy does not guarantee that its terminal state satisfies the preconditions of the next skill, particularly when navigation, bimanual manipulation~\cite{chen2026rotri}, and object transfer are interleaved.

\subsection{Harnessing VLMs for Embodied Task Planning}
A growing body of work structures the interface around a foundation model so that task decomposition produces executable behavior. Existing approaches constrain model outputs through affordance or feasibility filtering~\cite{ahn2022saycan}, program structure~\cite{liang2023code}, or formal planning and geometric checks~\cite{liu2023llmp,lin2023text2motion}. Embodied planners incorporate visual observations into sequential reasoning~\cite{driess2023palme,huang2023voxposer,chen2025owmm,chen2026deco}, whereas scene-graph systems plan over explicit environment representations~\cite{rana2023sayplan,ekpo2024verigraph,gu2024conceptgraphs}. Recent agentic systems additionally combine fine-tuned planning models with shared memory~\cite{tan2025roboos}, unified data-learning-execution frameworks~\cite{li2026roboclaw}, or persistent runtimes with critic feedback~\cite{huo2026abotclaw,chenadacleargrasp}. These designs make different trade-offs in model training, map maintenance, and memory infrastructure. Our focus is complementary: a frozen VLM is retained, while a compact, embodiment-specific interface supplies the state constraints and execution hooks.

\subsection{Closed-Loop Replanning and Failure Recovery}
Classical task and motion planning (TAMP)~\cite{garrett2021tamp} enforces symbolic--geometric consistency but depends on explicit domain models. Language-agent methods interleave reasoning with feedback~\cite{yao2022react,huang2022inner} or regenerate plans after failures~\cite{song2023llmplanner}. Other systems replan from perceptual feedback~\cite{skreta2024replan}, monitor plan-execution misalignment~\cite{guo2023doremi}, or reason about manipulation failures~\cite{duan2024aha,liu2025codeAsMonitor,chenadacleargrasp}. Frequent VLM queries can be costly over long horizons, and repairs that omit cumulative state can introduce new inconsistencies. In \pname, proprioceptive checks run continuously, semantic replanning is triggered only when an outcome invalidates the remaining plan, and every repaired sequence is subjected to the same SSG verification.

\section{METHOD}
\label{sec:method}

\begin{figure*}
    \vspace{7pt}
    \centering
    \includegraphics[width=0.93\linewidth]{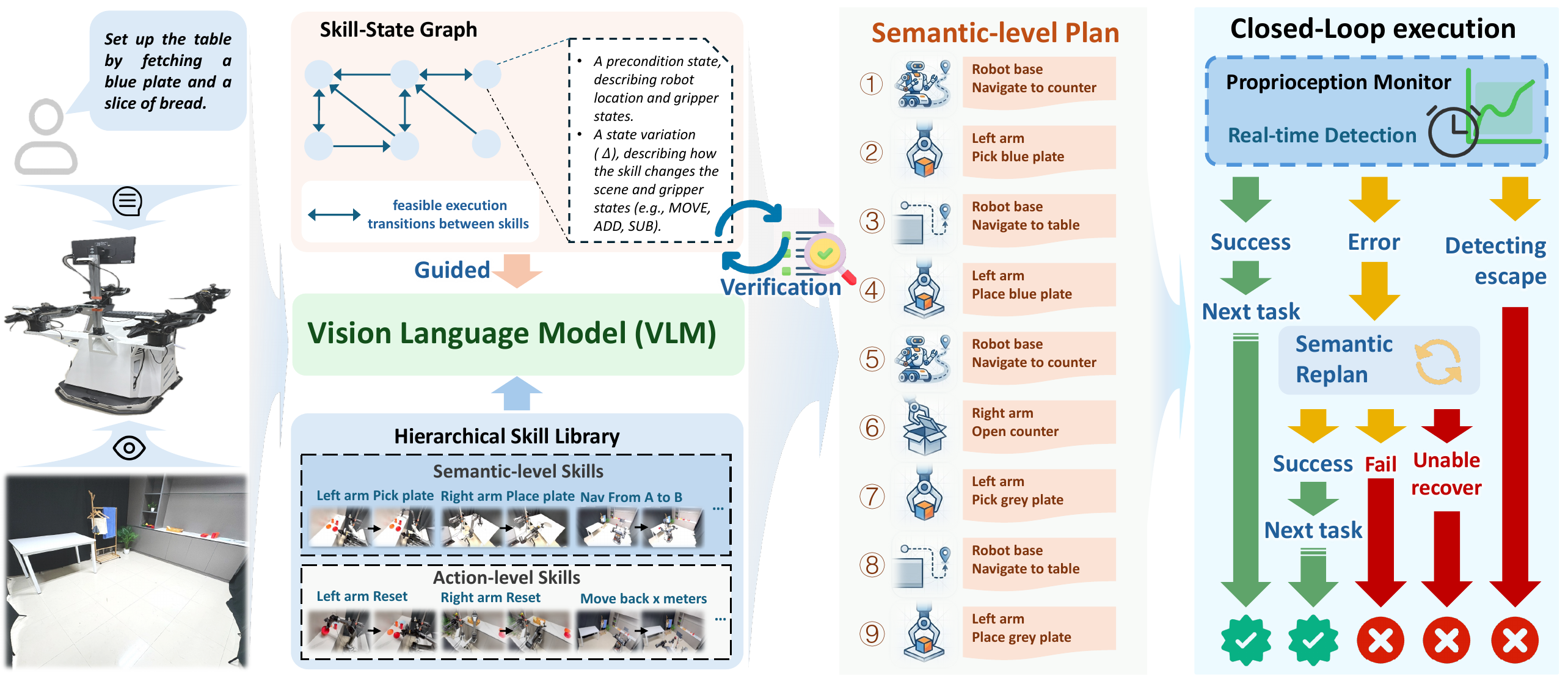}
    \caption{\small \textbf{Overview of \pname.} \pname harnesses a frozen VLM through three coupled mechanisms. (a) \emph{Constrain the interface}: the VLM decomposes an instruction over a topology-only projection of the Skill-State Graph (SSG), whose vocabulary is defined by a hierarchical skill library. (b) \emph{Verify before execution}: an SSG-based verifier simulates the cumulative robot-state trace of the proposed plan and returns structured errors for repair. (c) \emph{Query on events, not steps}: an event-driven monitor triggers graph-grounded repair when a skill outcome invalidates the remaining plan.}
    \label{fig:framework}
\vspace{-2em}
\end{figure*}

\pname receives a natural-language instruction $I$, visual observations $O_{\mathrm{vis}}$, and proprioceptive state $O_{\mathrm{ego}}$, and produces an executable semantic skill sequence $\mathbf{s}=[s_1,\ldots,s_K]$. The sequence is updated online only when execution diverges from an expected state transition. The setting is deliberately \emph{map-light}: the harness assumes named scene regions, a predefined skill library, and a manually specified skill-transition graph, but no dense metric map or complete symbolic world model. The VLM is frozen and serves as a proposal engine; embodiment-specific knowledge is encoded in the harness. As illustrated in Fig.~\ref{fig:framework}, the three mechanisms constrain the skill vocabulary and local transitions, verify cumulative robot-state feasibility before execution, and trigger graph-grounded repair only after an observed outcome invalidates the remaining plan.

\subsection{Hierarchical Skill Library and Skill-State Graph}
\label{sec:ssg}
The skill library has two levels: scene-semantics-aware \emph{semantic skills} and action-primitive \emph{action skills}. Semantic skills carry concrete task semantics, such as \textit{left-arm-pick-blue-plate}, \textit{navigate-from-dining-table-to-counter}, and \textit{right-arm-open-counter}, and are instantiated by imitation-learning policies, VLA policies, or rule-based controllers. Because they impose embodiment-state constraints, consecutive semantic skills must respect the state relations they induce. Action skills are atomic motor primitives, such as \textit{open-gripper}, \textit{close-gripper}, \textit{retract-arm-to-home}, and \textit{rotate-base-in-place}. They are decoupled from scene and task semantics, but remain interpretable to the VLM and executable by the underlying controllers. We expose these primitives by name rather than asking the VLM to emit joint-level commands. Their task-level preconditions are intentionally minimal, allowing them to bridge mismatched terminal and initial states or support recovery from a failed semantic skill. This separation reflects the fact that language models can propose high-level actions but do not directly provide reliable joint-level control without robot-action training~\cite{huang2022language,brohan2023rt2,kim2024openvla}.

The Skill-State Graph (SSG) makes these chaining constraints explicit by describing feasible transitions among semantic skills together with their embodiment-state requirements. For our dual-arm mobile manipulator, a robot state is
\begin{equation}
c=(L,h^l,h^r)\in\mathcal{C},
\label{eq:core_state}
\end{equation}
where $L$ is the discrete scene region occupied by the robot, $h^l$ and $h^r$ denote left- and right-gripper occupancy, and $\mathcal{C}$ is the set of admissible robot states. Formally, the SSG is a directed graph
\begin{equation}
\mathcal{G}=(\mathcal{V},\mathcal{E},\mathcal{P},\Delta),
\label{eq:ssg}
\end{equation}
whose nodes $v_k\in\mathcal{V}$ correspond to semantic skills $s_k$, whose directed edges $(v_k,v_l)\in\mathcal{E}$ denote locally feasible transitions, and whose nodes carry a \emph{precondition predicate} $\mathrm{pre}(s_k)\in\mathcal{P}$ and a \emph{state variation} $\Delta(s_k)$.

The precondition predicate $\mathrm{pre}(s_k)$ is a predicate over $\mathcal{C}$ that specifies the minimal robot-state requirements for executing $s_k$. For example, \textit{left-arm-pick-blue-plate} requires $L$ to be the region containing the blue plate and the left gripper to be free, whereas \textit{navigate-from-dining-table-to-counter} requires $L$ to be the dining-table region and imposes no gripper constraint. The state variation $\Delta(s_k):\mathcal{C}\rightarrow\mathcal{C}$ maps the pre-execution state to the state expected after successful completion of $s_k$ and is decomposed as
\begin{equation}
\Delta(s_k)=(f_L,f^l,f^r),
\label{eq:state_delta}
\end{equation}
where $f_L$ may apply $\mathrm{MOVE}(r_a,r_b)$ to the scene region and $f^l,f^r$ may apply $\mathrm{ADD}(o)$, $\mathrm{SUB}(o)$, or $\emptyset$ to each gripper. Thus, \textit{left-arm-pick-blue-plate} applies $\mathrm{ADD}(\textit{blue-plate})$ to $h^l$ with $L$ unchanged, whereas \textit{navigate-from-dining-table-to-counter} applies $\mathrm{MOVE}(\textit{dining-table},\textit{counter})$ and leaves both grippers unchanged. The SSG therefore captures the compact cumulative state needed to detect ordering violations while keeping graph maintenance and prompt serialization lightweight. Finer conditions, including receptacle state, support relations, and reachability, remain within the skills and their controllers.

\subsection{Graph-Guided Planning and Verification}
\label{sec:planning}
Given a long-horizon instruction $I$ and visual observation $O_{\mathrm{vis}}$, this stage generates a skill sequence $\mathbf{s}=[s_1,\ldots,s_K]$ through topology-aware semantic planning followed by cumulative robot-state verification.
In the planning stage, task decomposition is cast as routing over skill space, with the VLM constrained to transitions represented in the SSG. The VLM receives $I$, $O_{\mathrm{vis}}$, and a topological subgraph $\mathcal{G}_{\mathrm{topo}}\subset\mathcal{G}$ containing semantic-skill nodes and their adjacency but omitting $\mathcal{P}$ and $\Delta$. This compact representation specifies the available skill vocabulary and connectivity without exposing a dense scene model or symbolic object state. The VLM proposes a candidate $\hat{\mathbf{s}}=[\hat{s}_1,\ldots,\hat{s}_K]$ as a path over $\mathcal{G}_{\mathrm{topo}}$.

Topological validity does not imply executability: a path that is admissible at every edge can become infeasible as robot state accumulates. The verifier therefore replays the candidate with the full SSG attributes, starting from the initial state $c_0$ obtained from $O_{\mathrm{ego}}$ and recursively applying each state variation. The plan is accepted only when every skill is executable in the state it encounters:
\begin{equation}
c_{i-1}\models\mathrm{pre}(\hat{s}_i), \quad
c_i=\Delta(\hat{s}_i)(c_{i-1}),\;\;\forall i\in\{1,\ldots,K\},
\label{eq:verification}
\end{equation}
where $c\models\mathrm{pre}(s)$ denotes that the scene region $L$ and gripper occupancies $h^l,h^r$ in $c$ satisfy the precondition of $s$. When a conflict is detected, the verifier rejects the candidate and returns a structured error identifying the step and offending variable, for example, \textit{left gripper occupied before left-arm-pick-blue-plate}. The VLM repairs the conflicting portion, and the revised sequence is re-verified before execution. Thus, a $K$-step candidate incurs one precondition check per step in a verification pass, while prompt size scales with the extracted subgraph rather than the full library (Sec.~\ref{subsec:ablation}). Porting the state tuple in Eq.~\eqref{eq:core_state} requires redefining the regions, grippers, $\mathrm{pre}$, and $\Delta$ for the target embodiment.

\subsection{Closed-Loop Execution and Replanning}
\label{sec:execution}
\pname executes the validated sequence in a closed loop that combines real-time monitoring with graph-constrained recovery. For each skill $s_i$, \emph{ego-state monitoring} tracks proprioceptive signals such as joint encoders, gripper width, and tactile feedback to confirm primitive outcomes, such as a successful $\mathrm{ADD}$ or $\mathrm{SUB}$, without invoking the VLM. \emph{Semantic verification} is event-driven: a VLM scene check against the expected variation $\Delta(s_i)$ is invoked only when ego-state signals cannot confirm the outcome or when the skill changes the scene. If the observed state $c_{\mathrm{obs}}$ is consistent with the expected transition, execution proceeds to $s_{i+1}$; otherwise, a deviation is flagged.

Upon a deviation, such as a failed grasp or blocked navigation path, \pname replans over the SSG together with the action skills instead of restarting the task. With $c_{\mathrm{obs}}$ as the new initial state, a VLM re-query identifies a corrective path to a state satisfying the preconditions of the remaining subtasks while preserving completed ones. Action skills can be inserted along this path to compensate for the deviation, for example by reopening the gripper and regrasping, releasing an unexpectedly held object, or retracting the arm before re-navigation. Each corrective sequence is checked with Eq.~\eqref{eq:verification}. Deviations that cannot be represented by the state tuple, such as an object leaving the workspace, are reported as unrecoverable.

\section{EXPERIMENTS}
\label{sec:experiments}

We organize the evaluation around three research questions:
\textbf{RQ1 (Simulation effectiveness):} Does \pname improve planning validity and execution survival across tasks and skill configurations?
\textbf{RQ2 (Interface analysis):} How do backbone choice and the proposed interface components affect planning validity and computational efficiency, and what execution failures remain?
\textbf{RQ3 (Real-world transfer):} Does \pname retain its planning and recovery capabilities on a physical mobile manipulator across task difficulty levels?
Section~\ref{subsec:sim_results} addresses RQ1, Section~\ref{subsec:ablation} addresses RQ2, and Section~\ref{subsec:real_world_results} addresses RQ3.

\subsection{Simulation Effectiveness (RQ1)}
\label{subsec:sim_results}

\textbf{Simulation setup.} We use \textit{mshab*}, an extension of ManiSkill-Hab~\cite{shukla2024maniskillhab} with a VLM interface for task decomposition, API translation, skill chaining, and closed-loop recovery. The tasks are \textit{tidy\_house} (20 subtasks), \textit{prepare\_groceries} (12), and \textit{set\_table} (16). Navigation and open/close skills use pretrained PPO~\cite{schulman2017proximal}, while pick/place skills use pretrained SAC~\cite{haarnoja2018soft} in the RL configurations and behavioral cloning (BC)~\cite{bain1995framework} in the IL configuration. Each scene--configuration pair induces an SSG. Object-agnostic configurations (RL:All\_Obj and IL:All\_Obj) yield compact graphs with 3-7 skills and 4-12 edges, whereas the per-object RL:Per\_Obj configuration expands the graph to at most 19 skills and 36 edges. For the main experiments, we run three independent groups of 100 episodes for each task--SSG--method configuration, yielding 10,800 trajectories. The ablation studies include an additional 4,500 trajectories under the corresponding configurations. We report the mean and standard deviation across the three groups.

\textbf{Metrics.} Planning validity measures the fraction of generated plans that exactly match the ground-truth task sequence. Execution survival at step $k$ is the fraction of episodes that successfully complete the first $k$ subtasks. 

\textbf{Baselines.} RQ1 uses a frozen Qwen3.6 Flash~\cite{qwen2026qwen36flash} backbone. All methods receive the same instruction, observations, and underlying skill implementations, but differ in how skills are represented, composed, and checked. \textit{VLM-Direct} generates an open-loop sequence from a flat skill list without verification. \textit{DeCo*}~\cite{chen2026deco} is adapted to mobile manipulation and concatenates subgoal fragments without cumulative state checking. \textit{SayPlan*}~\cite{rana2023sayplan} retains scene-graph-grounded iterative planning with inputs and outputs adapted to our setting. For execution evaluation, \textit{Ground Truth (GT) Sequencing} provides a human-authored open-loop reference.

\begin{table*}
\vspace{7pt}
\centering
\caption{\small Planning validity in simulation (mean $\pm$ standard deviation, \%)}
\label{tab:combined_results}
\resizebox{\textwidth}{!}{
\begin{tabular}{c|ccc|ccc|ccc}
\hline
\multirow{2}{*}{\textbf{Methods}} & \multicolumn{3}{c|}{\textbf{Prepare\_Groceries (12 subtasks)}} & \multicolumn{3}{c|}{\textbf{Set\_Table (16 subtasks)}} & \multicolumn{3}{c}{\textbf{Tidy\_House (20 subtasks)}} \\
\cline{2-10}
 & \textbf{RL:Per\_Obj} & \textbf{RL:All\_Obj} & \textbf{IL:All\_Obj} & \textbf{RL:Per\_Obj} & \textbf{RL:All\_Obj} & \textbf{IL:All\_Obj} & \textbf{RL:Per\_Obj} & \textbf{RL:All\_Obj} & \textbf{IL:All\_Obj} \\
\hline
VLM-Direct & $1.7 {\scriptstyle \pm 1.2}$ & $26.3 {\scriptstyle \pm 1.2}$ & $32.3 {\scriptstyle \pm 2.9}$ & $6.0 {\scriptstyle \pm 1.7}$ & $12.3 {\scriptstyle \pm 1.2}$ & $10.3 {\scriptstyle \pm 1.2}$ & $1.0 {\scriptstyle \pm 1.0}$ & $22.3 {\scriptstyle \pm 2.5}$ & $17.7 {\scriptstyle \pm 1.5}$ \\
DeCo* & $3.0 {\scriptstyle \pm 3.5}$ & $31.7 {\scriptstyle \pm 2.3}$ & $27.7 {\scriptstyle \pm 2.3}$ & $7.3 {\scriptstyle \pm 0.6}$ & $11.3 {\scriptstyle \pm 0.6}$ & $12.7 {\scriptstyle \pm 0.6}$ & $3.7 {\scriptstyle \pm 1.5}$ & $25.0 {\scriptstyle \pm 2.0}$ & $23.0 {\scriptstyle \pm 1.0}$ \\
SayPlan* & $31.3 {\scriptstyle \pm 1.2}$ & $35.3 {\scriptstyle \pm 2.3}$ & $33.7 {\scriptstyle \pm 1.2}$ & $19.3 {\scriptstyle \pm 1.2}$ & $17.3 {\scriptstyle \pm 1.2}$ & $18.0 {\scriptstyle \pm 1.7}$ & $23.7 {\scriptstyle \pm 1.5}$ & $16.0 {\scriptstyle \pm 1.7}$ & $21.7 {\scriptstyle \pm 1.2}$ \\
\rowcolor{blue!10}
\textbf{MoMaStage} & $\mathbf{100.0} {\scriptstyle \pm 0.0}$ & $\mathbf{98.3} {\scriptstyle \pm 1.5}$ & $\mathbf{99.3} {\scriptstyle \pm 0.6}$ & $\mathbf{96.0} {\scriptstyle \pm 2.7}$ & $\mathbf{96.7} {\scriptstyle \pm 3.1}$ & $\mathbf{96.7} {\scriptstyle \pm 2.1}$ & $\mathbf{100.0} {\scriptstyle \pm 0.0}$ & $\mathbf{96.3} {\scriptstyle \pm 1.5}$ & $\mathbf{98.3} {\scriptstyle \pm 1.5}$ \\
\hline
\end{tabular}
}
\vspace{-1em}
\end{table*}

\begin{table*}
\centering
\caption{\small Execution survival at mid-horizon milestones in simulation (mean $\pm$ standard deviation, \%)}
\label{tab:consolidated_long}
\begin{adjustbox}{max width=\textwidth}
\begin{tabular}{c|c|ccc|ccc|cccc}
\hline
\multirow{2}{*}{\shortstack{SSG Type}} & \multirow{2}{*}{\text{Methods}} & \multicolumn{3}{c|}{Prepare\_Groceries } & \multicolumn{3}{c|}{Set\_Table} & \multicolumn{4}{c}{Tidy\_House} \\
\cline{3-12}
& & Step 4 & Step 6 & Step 8 & Step 4 & Step 6 & Step 8 & Step 4 & Step 8 & Step 12 & Step 16\\
\hline
\rowcolor{gray!20}
\cellcolor{white} & GT & 20.0\,${\scriptstyle\pm 2.0}$ & 12.0\,${\scriptstyle\pm 2.0}$ & 2.3\,${\scriptstyle\pm 1.5}$ & 82.3\,${\scriptstyle\pm 3.1}$ & 44.7\,${\scriptstyle\pm 1.1}$ & 25.0\,${\scriptstyle\pm 2.6}$ & 41.3\,${\scriptstyle\pm 3.5}$ & 15.0\,${\scriptstyle\pm 2.6}$ & 3.7\,${\scriptstyle\pm 1.1}$ & 0.7\,${\scriptstyle\pm 1.1}$ \\
\cellcolor{white} & DeCo* & 2.0\,${\scriptstyle\pm 3.5}$ & 1.3\,${\scriptstyle\pm 2.3}$ & 0.0\,${\scriptstyle\pm 0.0}$ & 4.3\,${\scriptstyle\pm 2.5}$ & 2.0\,${\scriptstyle\pm 2.6}$ & 1.0\,${\scriptstyle\pm 1.7}$ & 1.3\,${\scriptstyle\pm 1.1}$ & 0.0\,${\scriptstyle\pm 0.0}$ & 0.0\,${\scriptstyle\pm 0.0}$ & 0.0\,${\scriptstyle\pm 0.0}$ \\
\cellcolor{white} & SayPlan* & 5.7\,${\scriptstyle\pm 2.1}$ & 3.0\,${\scriptstyle\pm 1.7}$ & 0.7\,${\scriptstyle\pm 0.6}$ & 14.7\,${\scriptstyle\pm 4.0}$ & 7.7\,${\scriptstyle\pm 3.5}$ & 3.7\,${\scriptstyle\pm 1.5}$ & 9.0\,${\scriptstyle\pm 4.4}$ & 1.7\,${\scriptstyle\pm 1.5}$ & 0.0\,${\scriptstyle\pm 0.0}$ & 0.0\,${\scriptstyle\pm 0.0}$ \\
\rowcolor{blue!10}
\cellcolor{white}\multirow{-4}{*}{\shortstack{RL:Per\_Obj \\ (SAC+PPO)}} & \textbf{MoMaStage} & \textbf{27.0}\,${\scriptstyle\pm 1.0}$ & \textbf{16.3}\,${\scriptstyle\pm 2.9}$ & \textbf{5.3}\,${\scriptstyle\pm 2.9}$ & \textbf{84.7}\,${\scriptstyle\pm 0.6}$ & \textbf{49.7}\,${\scriptstyle\pm 3.5}$ & \textbf{30.0}\,${\scriptstyle\pm 0.0}$ & \textbf{47.0}\,${\scriptstyle\pm 2.6}$ & \textbf{21.0}\,${\scriptstyle\pm 8.0}$ & \textbf{8.3}\,${\scriptstyle\pm 4.0}$ & \textbf{2.7}\,${\scriptstyle\pm 2.1}$ \\
\hline
\rowcolor{gray!20}
\cellcolor{white} & GT & 18.3\,${\scriptstyle\pm 4.2}$ & 5.7\,${\scriptstyle\pm 0.6}$ & 3.3\,${\scriptstyle\pm 0.6}$ & 72.7\,${\scriptstyle\pm 2.1}$ & 32.3\,${\scriptstyle\pm 2.5}$ & 22.0\,${\scriptstyle\pm 2.0}$ & 34.7\,${\scriptstyle\pm 2.5}$ & 10.0\,${\scriptstyle\pm 1.0}$ & 2.3\,${\scriptstyle\pm 0.6}$ & 0.3\,${\scriptstyle\pm 0.6}$ \\
\cellcolor{white} & DeCo* & 3.7\,${\scriptstyle\pm 1.5}$ & 0.0\,${\scriptstyle\pm 0.0}$ & 0.0\,${\scriptstyle\pm 0.0}$ & 10.0\,${\scriptstyle\pm 4.4}$ & 5.0\,${\scriptstyle\pm 3.6}$ & 2.3\,${\scriptstyle\pm 2.5}$ & 8.7\,${\scriptstyle\pm 5.5}$ & 0.7\,${\scriptstyle\pm 0.6}$ & 0.0\,${\scriptstyle\pm 0.0}$ & 0.0\,${\scriptstyle\pm 0.0}$ \\
\cellcolor{white} & SayPlan* & 6.0\,${\scriptstyle\pm 2.6}$ & 3.0\,${\scriptstyle\pm 2.0}$ & 0.7\,${\scriptstyle\pm 0.6}$ & 11.3\,${\scriptstyle\pm 4.0}$ & 7.7\,${\scriptstyle\pm 3.1}$ & 4.3\,${\scriptstyle\pm 1.1}$ & 8.3\,${\scriptstyle\pm 3.1}$ & 3.3\,${\scriptstyle\pm 0.6}$ & 0.0\,${\scriptstyle\pm 0.0}$ & 0.0\,${\scriptstyle\pm 0.0}$ \\
\rowcolor{blue!10}
\cellcolor{white}\multirow{-4}{*}{\shortstack{RL:All\_Obj \\ (SAC+PPO)}} & \textbf{MoMaStage} & \textbf{19.0}\,${\scriptstyle\pm 2.0}$ & \textbf{8.0}\,${\scriptstyle\pm 2.0}$ & \textbf{3.7}\,${\scriptstyle\pm 2.1}$ & \textbf{78.3}\,${\scriptstyle\pm 7.6}$ & \textbf{35.7}\,${\scriptstyle\pm 4.0}$ & \textbf{24.3}\,${\scriptstyle\pm 4.6}$ & 32.3\,${\scriptstyle\pm 4.0}$ & \textbf{12.3}\,${\scriptstyle\pm 2.3}$ & \textbf{4.0}\,${\scriptstyle\pm 0.0}$ & \textbf{1.7}\,${\scriptstyle\pm 1.1}$ \\
\hline
\rowcolor{gray!20}
\cellcolor{white} & GT & 11.7\,${\scriptstyle\pm 3.8}$ & 4.3\,${\scriptstyle\pm 1.5}$ & 1.0\,${\scriptstyle\pm 1.0}$ & 70.3\,${\scriptstyle\pm 3.2}$ & 38.3\,${\scriptstyle\pm 3.5}$ & 17.3\,${\scriptstyle\pm 2.1}$ & 38.7\,${\scriptstyle\pm 3.5}$ & 9.0\,${\scriptstyle\pm 0.0}$ & 3.0\,${\scriptstyle\pm 1.0}$ & 0.3\,${\scriptstyle\pm 0.6}$ \\
\cellcolor{white} & DeCo* & 3.0\,${\scriptstyle\pm 1.0}$ & 3.0\,${\scriptstyle\pm 1.0}$ & 0.0\,${\scriptstyle\pm 0.0}$ & 10.0\,${\scriptstyle\pm 1.0}$ & 4.3\,${\scriptstyle\pm 1.1}$ & 2.7\,${\scriptstyle\pm 0.6}$ & 8.7\,${\scriptstyle\pm 4.5}$ & 2.3\,${\scriptstyle\pm 1.5}$ & 0.3\,${\scriptstyle\pm 0.6}$ & 0.0\,${\scriptstyle\pm 0.0}$ \\
\cellcolor{white} & SayPlan* & 4.7\,${\scriptstyle\pm 0.6}$ & 3.7\,${\scriptstyle\pm 1.5}$ & 0.5\,${\scriptstyle\pm 0.7}$ & 13.3\,${\scriptstyle\pm 2.1}$ & 6.7\,${\scriptstyle\pm 1.5}$ & 4.7\,${\scriptstyle\pm 1.1}$ & 11.7\,${\scriptstyle\pm 4.2}$ & 3.3\,${\scriptstyle\pm 1.1}$ & 0.7\,${\scriptstyle\pm 1.1}$ & 0.0\,${\scriptstyle\pm 0.0}$ \\
\rowcolor{blue!10}
\cellcolor{white}\multirow{-4}{*}{\shortstack{IL:All\_Obj \\ (BC+PPO)}} & \textbf{MoMaStage} & \textbf{17.0}\,${\scriptstyle\pm 3.6}$ & \textbf{7.3}\,${\scriptstyle\pm 6.1}$ & \textbf{3.0}\,${\scriptstyle\pm 2.6}$ & \textbf{72.3}\,${\scriptstyle\pm 1.1}$ & \textbf{42.7}\,${\scriptstyle\pm 0.6}$ & \textbf{23.0}\,${\scriptstyle\pm 1.7}$ & \textbf{40.3}\,${\scriptstyle\pm 4.2}$ & \textbf{10.7}\,${\scriptstyle\pm 2.5}$ & \textbf{5.7}\,${\scriptstyle\pm 2.3}$ & \textbf{3.0}\,${\scriptstyle\pm 1.0}$ \\
\hline
\end{tabular}
\end{adjustbox}
\vspace{-2em}
\end{table*}

\textbf{Planning validity.} Table~\ref{tab:combined_results} compares the four methods across three tasks and skill configurations. \pname achieves 96.0\%-100.0\% validity in all conditions, whereas the baselines range from 1.0\% to 35.3\%. VLM-Direct is particularly unreliable on longer tasks and per-object configurations, consistent with the difficulty of tracking cumulative embodiment state from a flat skill list. DeCo* and SayPlan* improve over direct planning in some conditions but remain below \pname. This advantage persists across task horizons, object-specific and object-agnostic graphs, and RL- and IL-based skill implementations. The larger RL:Per\_Obj graphs do not uniformly reduce validity, suggesting a trade-off between graph compactness and the semantic cues provided by object-specific skills. More broadly, the results suggest that VLM planning benefits more from explicit \textbf{skill-to-skill dependencies} than from richer scene descriptions or additional subgoal decomposition alone. Exposing enabling, ordering, and constraint relations provides the structure needed to generate executable sequences, indicating that interface quality depends on the clarity of skill transitions rather than the amount of context. These results establish the system-level planning benefit of \pname. RQ2 subsequently isolates the contributions of topology extraction, SSG verification, and graph-grounded repair.

\textbf{Execution survival.} GT provides an oracle-authored open-loop reference, whereas \pname can repair observed deviations online. Table~\ref{tab:consolidated_long} reports representative mid-horizon milestones, chosen to expose how survival changes after planning and early execution. DeCo* and SayPlan* decline rapidly because their execution processes cannot compensate for invalid or failed preceding skills. GT remains competitive early because its initial sequence is human-authored, but its survival degrades as execution errors accumulate without repair. In contrast, \pname exceeds GT at most reported milestones, including later steps where repeated navigation and manipulation increase the probability of deviation. The gains are consistent with graph-grounded replanning restoring an executable state after a subset of failed skills rather than restarting the full task. Overall, RQ1 shows that \pname improves planning validity and extends execution survival across the evaluated tasks and skill configurations, although final completion remains limited by low-level execution failures.

\subsection{Interface Analysis and Remaining Failures (RQ2)}
\label{subsec:ablation}

RQ1 establishes the system-level benefits of \pname but does not identify whether they arise from a particular VLM or from the proposed interface. RQ2 therefore separates three factors: backbone portability and computational cost, topology-only SSG extraction, and verification-guided repair. It then analyzes the terminal failures that remain after these mechanisms are applied. Ablations use Qwen3.6 Flash~\cite{qwen2026qwen36flash}, Gemini 3.6 Flash~\cite{google2026gemini36flash}, and GPT-5.4 nano~\cite{openai2026gpt54nano}. \textit{MoMaStage*} receives the full SSG while retaining the same cumulative-state verification and repair procedure as \pname, thereby isolating topology-only extraction from the remaining framework.

\begin{table}
\centering
\caption{\small Computational ablation on \textit{Prepare\_Groceries} under the RL:Per\_Obj configuration. Tokens and Think. Tok. denote total and reasoning-token counts, respectively.}
\label{tab:performance_metrics}
\resizebox{\columnwidth}{!}{\begin{tabular}{ccccc}
\toprule
VLM & Method & Time (s) $\downarrow$ & Tokens $\downarrow$ & Think. Tok. $\downarrow$ \\
\midrule
& VLM-Direct & 36.7\,${\scriptstyle\pm 0.4}$ & 6145.7\,${\scriptstyle\pm 26.3}$ & 5264.7\,${\scriptstyle\pm 42.7}$ \\
& DeCo* & 38.2\,${\scriptstyle\pm 0.6}$ & 6246.3\,${\scriptstyle\pm 43.6}$ & 5316.0\,${\scriptstyle\pm 40.0}$ \\
& SayPlan* & 28.2\,${\scriptstyle\pm 0.5}$ & 5383.7\,${\scriptstyle\pm 30.7}$ & 4278.3\,${\scriptstyle\pm 19.6}$ \\
& MoMaStage* & 30.9\,${\scriptstyle\pm 0.6}$ & 7400.7\,${\scriptstyle\pm 29.5}$ & 4595.7\,${\scriptstyle\pm 28.7}$ \\
\rowcolor{blue!10}
\cellcolor{white}\multirow{-5}{*}{\shortstack{Qwen3.6\\Flash}} & \textbf{MoMaStage} & \textbf{26.3}\,${\scriptstyle\pm 0.6}$ & \textbf{6069.0}\,${\scriptstyle\pm 38.2}$ & \textbf{3891.0}\,${\scriptstyle\pm 38.7}$ \\
\midrule
& VLM-Direct & 9.7\,${\scriptstyle\pm 0.3}$ & 4287.7\,${\scriptstyle\pm 25.4}$ & -- \\
& DeCo* & 11.8\,${\scriptstyle\pm 0.3}$ & 4684.7\,${\scriptstyle\pm 44.5}$ & -- \\
& SayPlan* & 11.1\,${\scriptstyle\pm 0.5}$ & 4520.3\,${\scriptstyle\pm 34.9}$ & -- \\
& MoMaStage* & 10.6\,${\scriptstyle\pm 0.5}$ & 6040.3\,${\scriptstyle\pm 41.4}$ & -- \\
\rowcolor{blue!10}
\cellcolor{white}\multirow{-5}{*}{\shortstack{Gemini 3.6\\Flash}} & \textbf{MoMaStage} & \textbf{7.8}\,${\scriptstyle\pm 0.6}$ & 4841.0\,${\scriptstyle\pm 37.8}$ & -- \\
\bottomrule
\end{tabular}}
\vspace{-1em}
\end{table}

\begin{table}
\centering
\caption{\small Planning-validity ablation on \textit{Prepare\_Groceries} using GPT-5.4 nano~\cite{openai2026gpt54nano}. Initial and Final denote strict validity of the initial proposal and the final SSG-verified plan after graph-grounded repair, respectively.}
\label{tab:planning_metrics_simplified}
\begin{tabular}{cccc}
\toprule
SSG Type & Method & Initial Val. $\uparrow$ & Final Val. $\uparrow$ \\
\midrule
& SayPlan*       & 18.7\,${\scriptstyle\pm 1.5}$ & 18.7\,${\scriptstyle\pm 1.5}$ \\
& MoMaStage*     & 19.7\,${\scriptstyle\pm 1.5}$ & 98.0\,${\scriptstyle\pm 1.0}$ \\
\rowcolor{blue!10}
\cellcolor{white}\multirow{-3}{*}{\shortstack{RL:Per\_Obj\\(SAC+PPO)}} & \textbf{MoMaStage} & \textbf{38.3}\,${\scriptstyle\pm 1.2}$ & \textbf{100.0}\,${\scriptstyle\pm 0.0}$ \\
\midrule
& SayPlan*       & 20.3\,${\scriptstyle\pm 1.5}$ & 20.3\,${\scriptstyle\pm 1.5}$ \\
& MoMaStage*     & 43.3\,${\scriptstyle\pm 1.5}$ & 92.7\,${\scriptstyle\pm 0.6}$ \\
\rowcolor{blue!10}
\cellcolor{white}\multirow{-3}{*}{\shortstack{RL:All\_Obj\\(SAC+PPO)}} & \textbf{MoMaStage} & \textbf{75.0}\,${\scriptstyle\pm 2.0}$ & \textbf{98.7}\,${\scriptstyle\pm 0.6}$ \\
\midrule
& SayPlan*       & 18.7\,${\scriptstyle\pm 0.6}$ & 18.7\,${\scriptstyle\pm 0.6}$ \\
& MoMaStage*     & 31.7\,${\scriptstyle\pm 2.1}$ & 93.3\,${\scriptstyle\pm 0.6}$ \\
\rowcolor{blue!10}
\cellcolor{white}\multirow{-3}{*}{\shortstack{IL:All\_Obj\\(BC+PPO)}} & \textbf{MoMaStage} & \textbf{78.7}\,${\scriptstyle\pm 1.2}$ & \textbf{98.3}\,${\scriptstyle\pm 1.2}$ \\
\bottomrule
\end{tabular}
\vspace{-2em}
\end{table}

\textbf{Backbone portability and computational cost.} Absolute costs differ across VLM backbones and are therefore compared only within each backbone. Table~\ref{tab:performance_metrics} shows that \pname has the lowest latency on both Qwen3.6 Flash~\cite{qwen2026qwen36flash} and Gemini 3.6 Flash~\cite{google2026gemini36flash}. On Qwen, the only backbone exposing reasoning-token counts, it also uses the fewest reasoning tokens, while its total-token use remains comparable to the unverified baselines. Gemini does not expose reasoning-token counts, and total-token use is not uniformly lowest across all methods. Nevertheless, topology-only extraction consistently reduces both latency and total tokens relative to MoMaStage*. Thus, latency improvements appear across both backbones, whereas the reasoning-token result is limited to the available Qwen measurement.

\textbf{Topology-only SSG extraction.} MoMaStage* receives the full SSG, whereas \pname exposes only the task-relevant topology. Across both backbones in Table~\ref{tab:performance_metrics}, the focused representation lowers latency and token use relative to MoMaStage*. Table~\ref{tab:planning_metrics_simplified} shows that it also improves first-attempt validity under every GPT-5.4 nano~\cite{openai2026gpt54nano} configuration. The magnitude varies with graph and skill-policy configuration, but the direction is consistent: removing instruction-irrelevant transitions reduces prompt complexity while preserving the connectivity needed for planning. The comparison therefore attributes the initial-validity and efficiency improvements to focused graph exposure rather than to SSG verification alone.

\textbf{SSG verification and graph-grounded repair.} The Initial and Final columns in Table~\ref{tab:planning_metrics_simplified} separate proposal quality from verification-guided repair. SayPlan* shows no improvement because its scene-level representation does not encode the cumulative skill preconditions checked by the SSG. In contrast, \pname and MoMaStage* raise low initial validity to 92.7\%--100.0\% final validity across all graph configurations. This result is especially informative with GPT-5.4 nano~\cite{openai2026gpt54nano}: even when the initial proposal is frequently invalid, structured feedback identifies the conflicting step and produces a sequence satisfying the encoded transitions and cumulative preconditions. Together, the comparisons show that topology extraction improves initial proposals, while SSG verification and graph-grounded repair drive the increase in final validity.

\begin{table}
\vspace{7pt}
\centering
\caption{\small \textbf{Distribution of terminal failure modes in simulation (\%)}. FLE: Force Limit Exceeded; TLE: Time Limit Exceeded; PTF: Preceding Task Failed.}
\label{tab:failure_modes}
\footnotesize
\setlength{\tabcolsep}{4pt}
\begin{tabular}{ccccc}
\toprule
\multicolumn{1}{c}{\textbf{Scene}} & \multicolumn{1}{c}{\textbf{Graph}} & \multicolumn{1}{c}{\textbf{FLE}} & \multicolumn{1}{c}{\textbf{TLE}} & \multicolumn{1}{c}{\textbf{PTF}} \\
\midrule
\multirow{3}{*}{\textit{Prepare\_Groceries}}
& RL:Per\_Obj & 52.3 ${\scriptstyle \pm 5.0}$ & 47.7 ${\scriptstyle \pm 5.0}$ & 0.0 ${\scriptstyle \pm 0.0}$ \\
& RL:All\_Obj & 59.3 ${\scriptstyle \pm 4.5}$ & 40.7 ${\scriptstyle \pm 4.5}$ & 0.0 ${\scriptstyle \pm 0.0}$ \\
& IL:All\_Obj & 52.0 ${\scriptstyle \pm 5.6}$ & 48.0 ${\scriptstyle \pm 5.6}$ & 0.0 ${\scriptstyle \pm 0.0}$ \\
\midrule
\multirow{3}{*}{\textit{Set\_Table}}
& RL:Per\_Obj & 70.7 ${\scriptstyle \pm 4.7}$ & 29.3 ${\scriptstyle \pm 4.7}$ & 0.0 ${\scriptstyle \pm 0.0}$ \\
& RL:All\_Obj & 70.7 ${\scriptstyle \pm 1.5}$ & 29.3 ${\scriptstyle \pm 1.5}$ & 0.0 ${\scriptstyle \pm 0.0}$ \\
& IL:All\_Obj & 72.7 ${\scriptstyle \pm 3.5}$ & 27.3 ${\scriptstyle \pm 3.5}$ & 0.0 ${\scriptstyle \pm 0.0}$ \\
\midrule
\multirow{3}{*}{\textit{Tidy\_House}}
& RL:Per\_Obj & 80.0 ${\scriptstyle \pm 5.0}$ & 20.0 ${\scriptstyle \pm 5.0}$ & 0.0 ${\scriptstyle \pm 0.0}$ \\
& RL:All\_Obj & 81.0 ${\scriptstyle \pm 3.0}$ & 19.0 ${\scriptstyle \pm 3.0}$ & 0.0 ${\scriptstyle \pm 0.0}$ \\
& IL:All\_Obj & 83.7 ${\scriptstyle \pm 2.5}$ & 16.3 ${\scriptstyle \pm 2.5}$ & 0.0 ${\scriptstyle \pm 0.0}$ \\
\bottomrule
\end{tabular}
\vspace{-2em}
\end{table}

\textbf{Remaining failure modes.} Table~\ref{tab:failure_modes} characterizes terminal failures remaining after planning verification and online repair. FLE dominates and is associated with simulator-level contact anomalies such as mesh clipping during manipulation. Under our protocol, FLE is treated as an unrecoverable physical error: the episode terminates immediately and no replanning is issued. TLE indicates that a controller exceeded its time budget, while PTF indicates that a downstream skill was blocked by an unsuccessful preceding skill. Both are treated as potentially recoverable deviations and trigger replanning; they become terminal only when recovery cannot restore an executable state. This distinction separates planning validity from physical completion: SSG verification addresses encoded state inconsistencies, whereas contact, grasping, navigation, and controller failures remain governed by the underlying skills. Overall, RQ2 shows that topology extraction improves proposal quality and efficiency, verification and repair drive final validity, and physical execution remains bounded by failures outside the compact SSG state.

\begin{figure}
    \vspace{1em}
    \centering
    \captionsetup[subfigure]{font=small}
    \begin{minipage}{0.96\linewidth}
        \begin{subfigure}{0.48\linewidth}
            \centering
            \includegraphics[width=\linewidth]{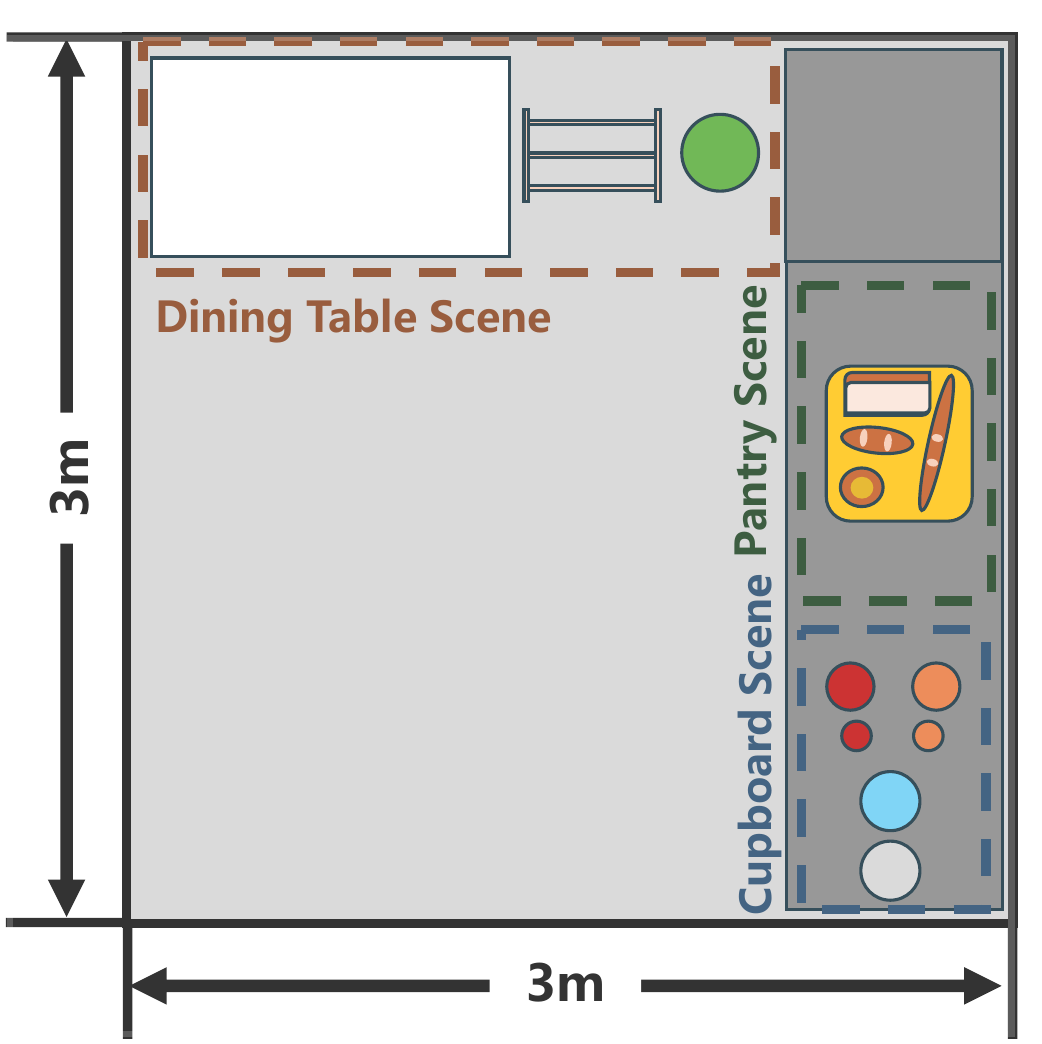}
            \caption{Top-down schematic of the scene layout.}
            \label{fig:MoMaStage_realrobot_scene}
        \end{subfigure}\hfill
        \begin{subfigure}{0.48\linewidth}
            \centering
            \includegraphics[width=\linewidth]{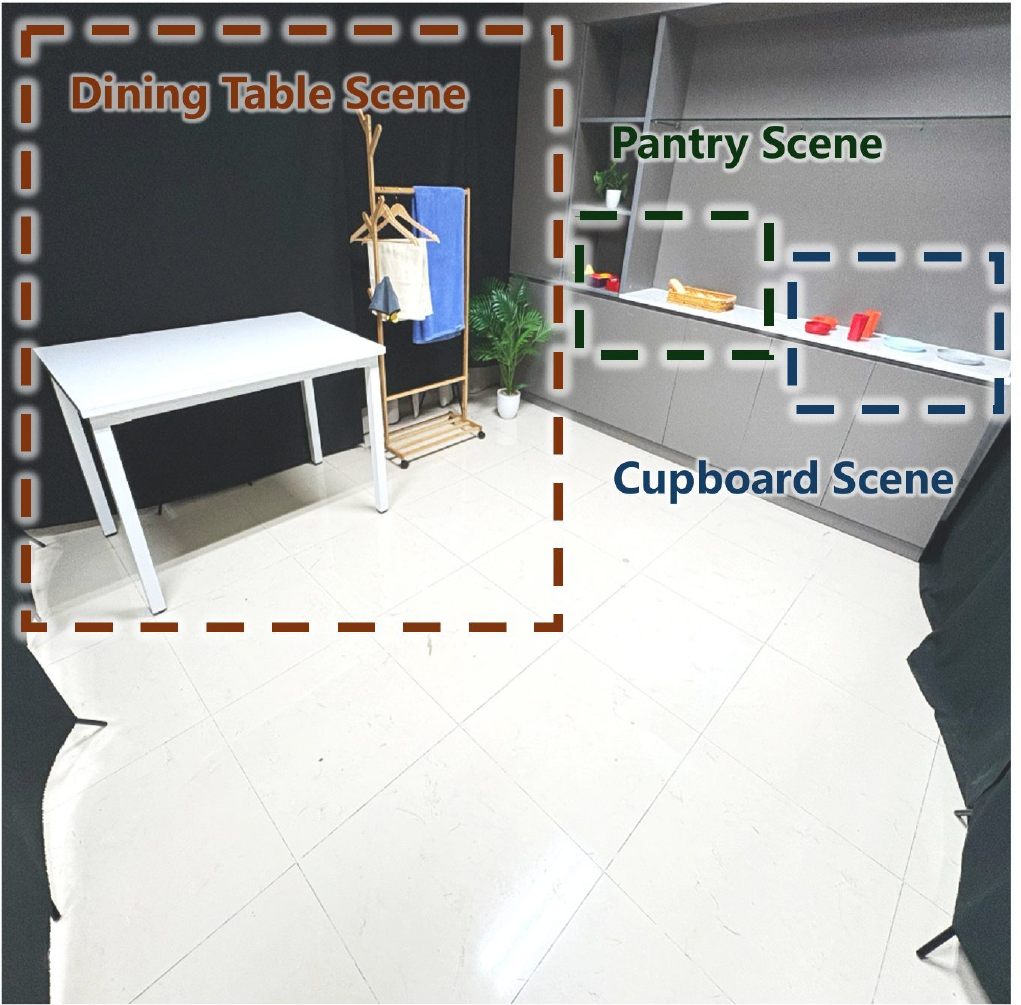}
            \caption{Photograph of the scene configuration.}
            \label{fig:MoMaStage_realrobot_RealSetup}
        \end{subfigure}

        \vspace{0.3cm}

        \begin{subfigure}{0.48\linewidth}
            \centering
            \includegraphics[width=\linewidth]{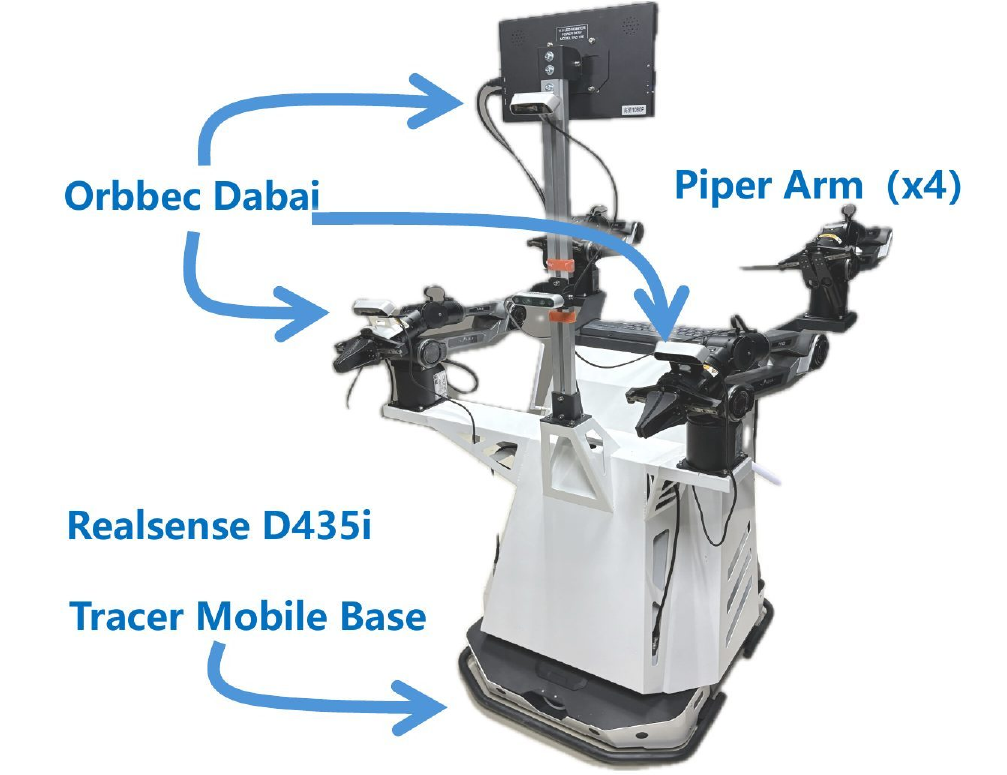}
            \caption{Hardware configuration of the robotic platform.}
            \label{fig:MoMaStage_realrobot_CobotMagic}
        \end{subfigure}\hfill
        \begin{subfigure}{0.48\linewidth}
            \centering
            \includegraphics[width=\linewidth]{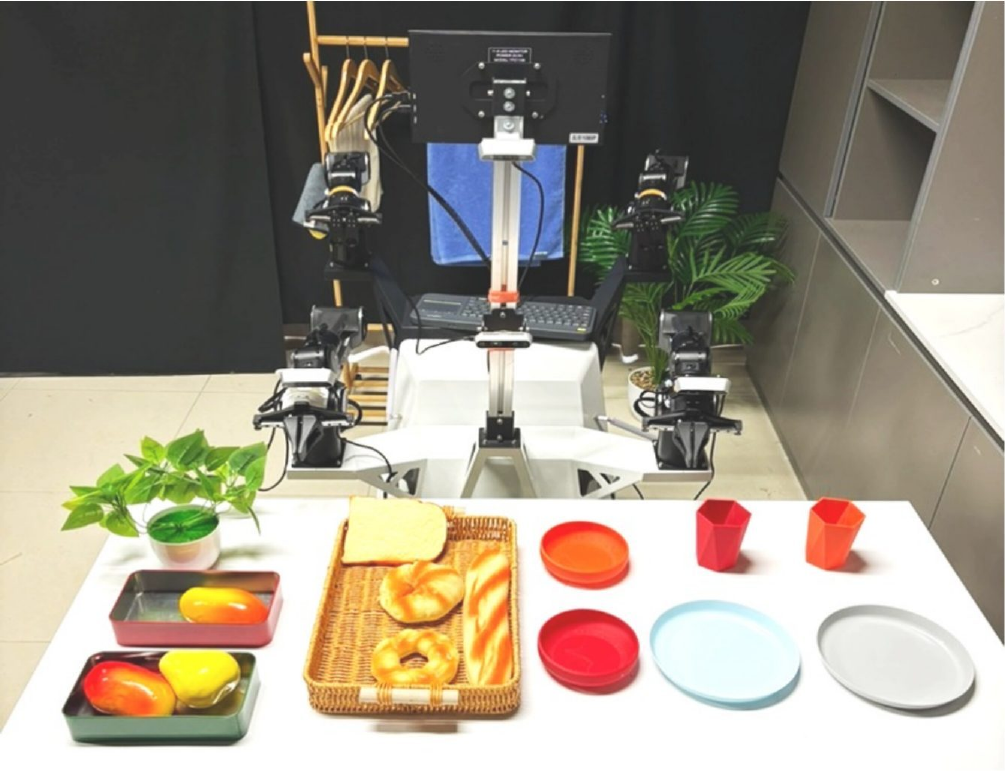}
            \caption{Objects and furniture used in the experiments.}
            \label{fig:MoMaStage_realrobot_Object}
        \end{subfigure}
    \end{minipage}
    \caption{\small \textbf{Overview of the real-world experimental setup.}}
    \label{fig:MoMaStage_realworld_overview}
\vspace{-2em}
\end{figure}

\begin{figure*}
\vspace{7pt}
    \centering
    \captionsetup[subfigure]{font=small}
    \begin{minipage}{0.93\textwidth}
        \begin{minipage}{0.35\linewidth}
            \begin{subfigure}{\linewidth}
                \centering
                \includegraphics[width=\linewidth]{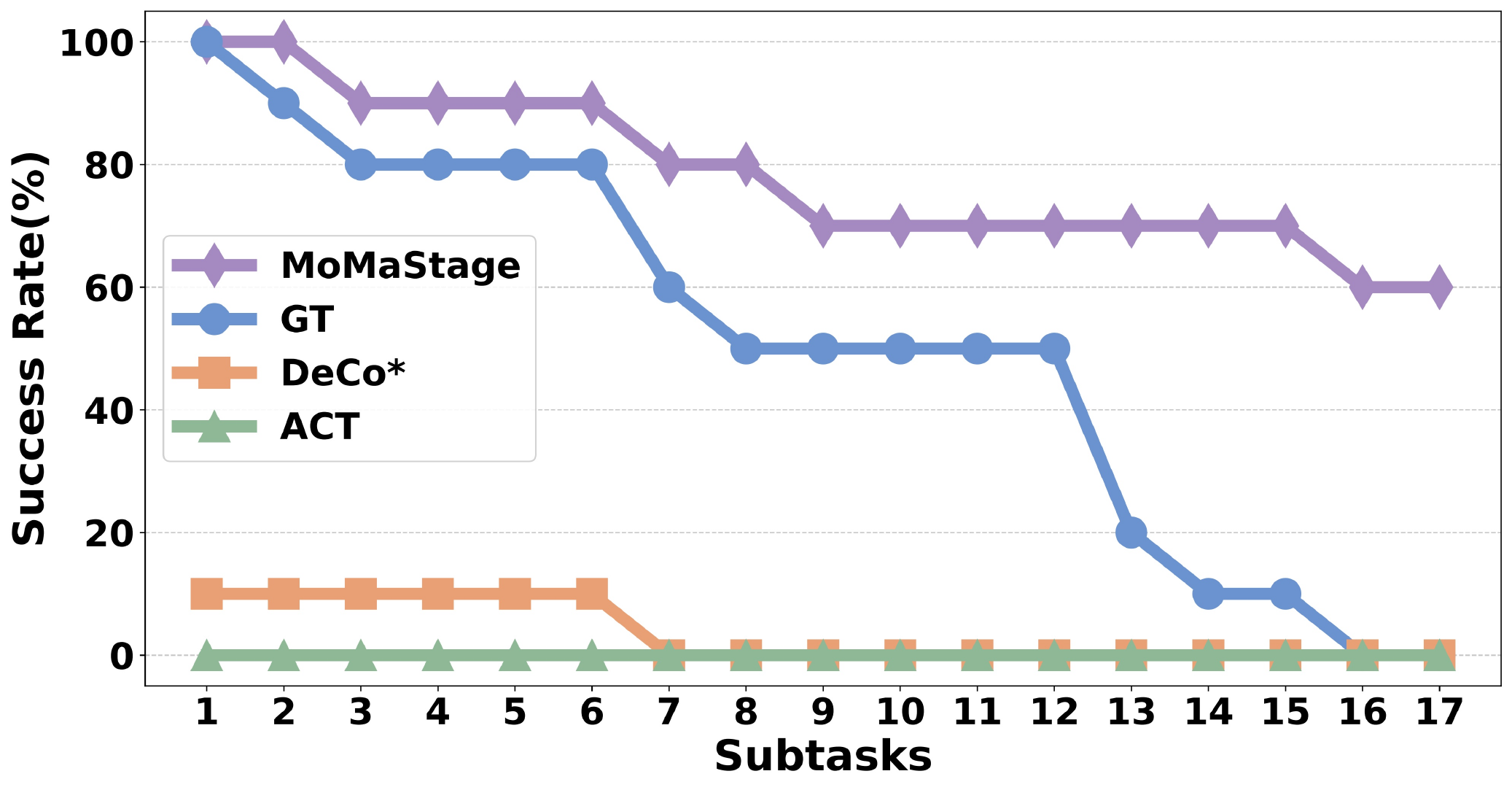}
                \caption{Per-subtask success rates.}
                \label{fig:MoMaStage_realrobot_sr}
            \end{subfigure}
            \vspace{0.2cm}
            \begin{subfigure}{\linewidth}
                \centering
                \includegraphics[width=\linewidth]{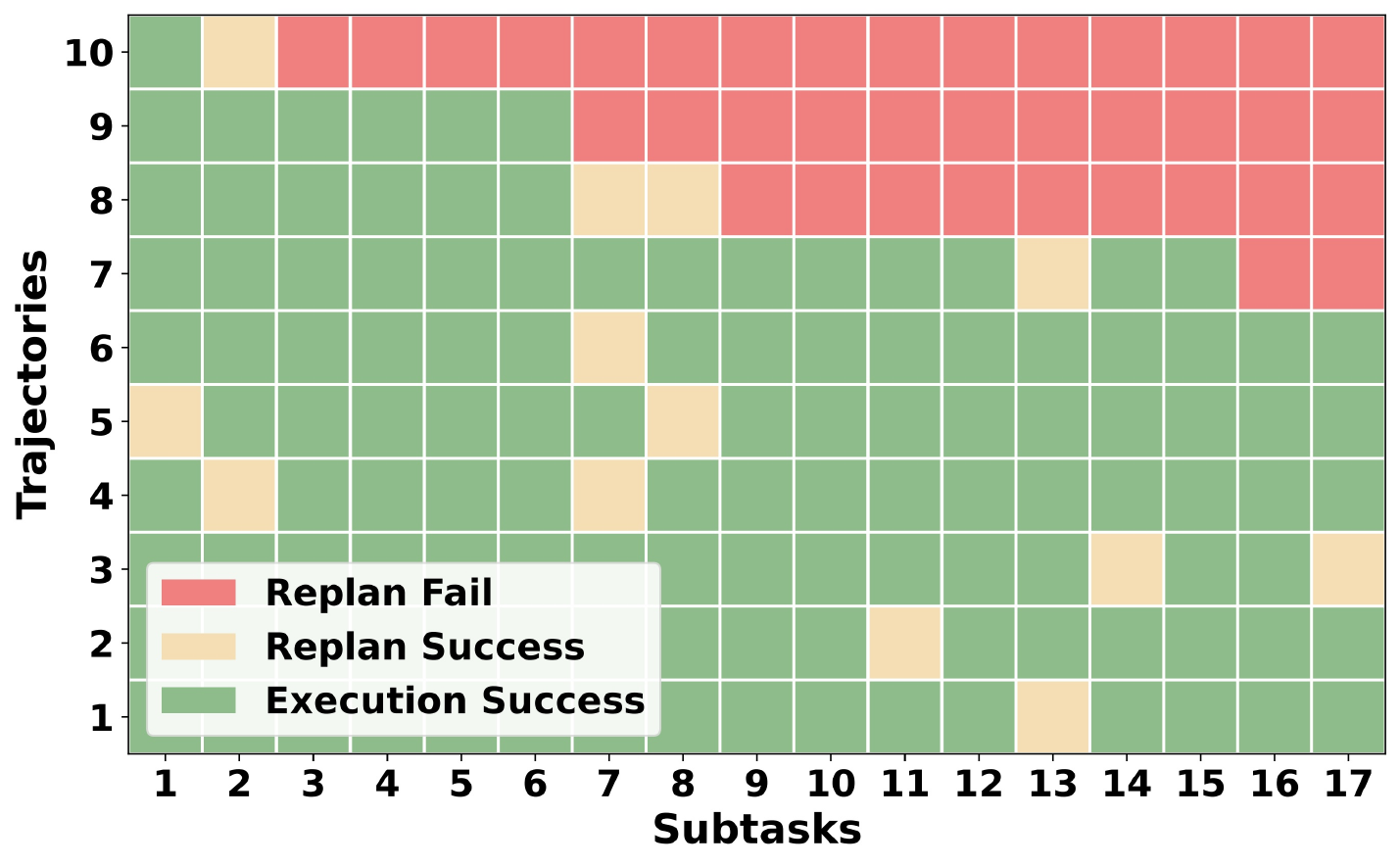}
                \caption{Detailed execution states of 10 trials.}
                \label{fig:MoMaStage_realrobot_detail}
            \end{subfigure}
        \end{minipage}\hfill
        \begin{minipage}{0.63\linewidth}
            \begin{subfigure}{\linewidth}
                \centering
                \includegraphics[width=\linewidth]{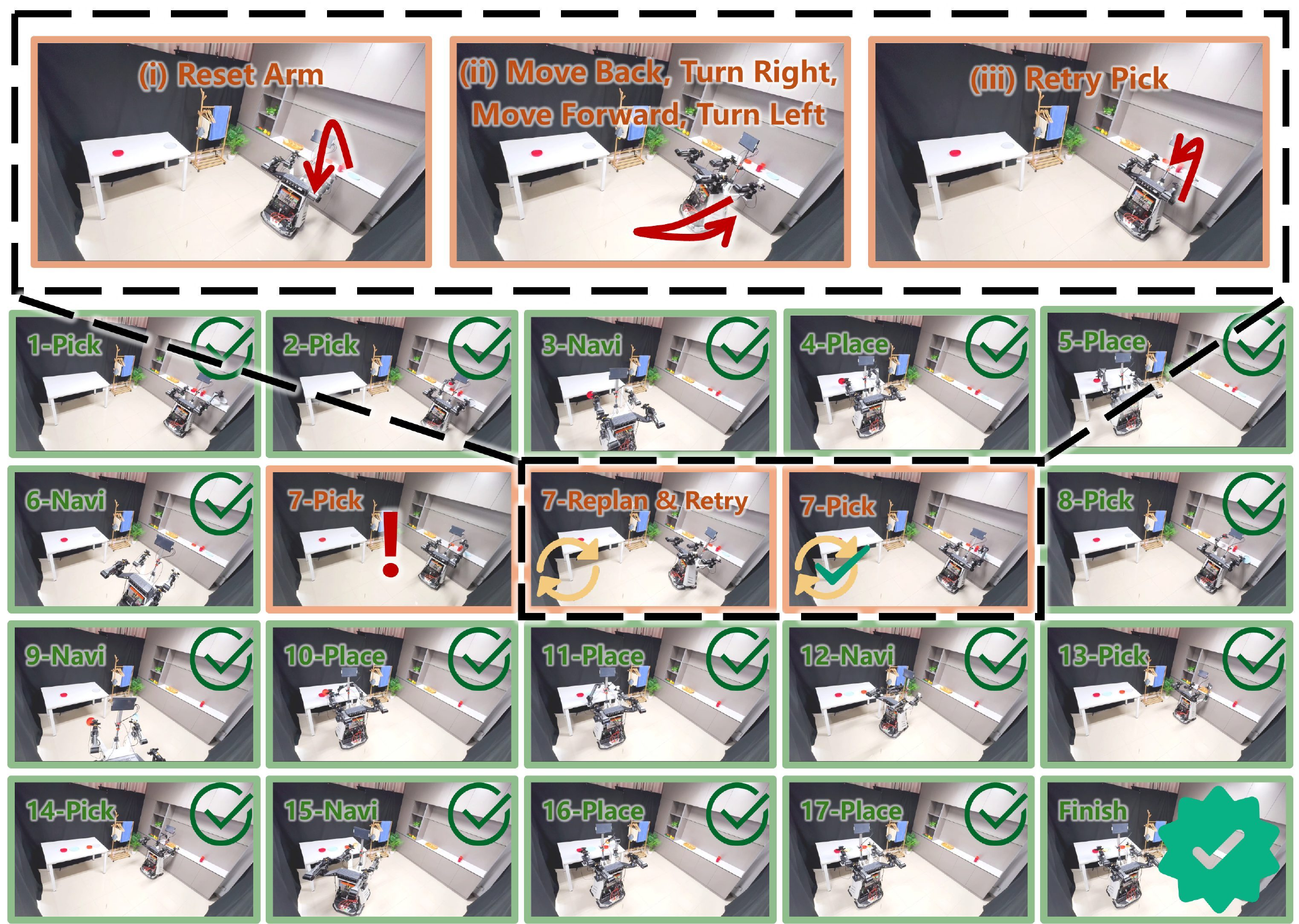}
                \caption{Successful recovery from a failure during subtask 6.}
                \label{fig:MoMaStage_realrobot_traj1show}
            \end{subfigure}
        \end{minipage}
    \end{minipage}
    \caption{\small \textbf{Evaluation of MoMaStage on the real-world robot.} (a) Per-subtask success compared with the baselines. (b) Execution states for 10 trials. (c) A long-horizon trajectory with event-triggered replanning.}
    \label{fig:MoMaStage_realrobot_overall}
\vspace{-2em}
\end{figure*}

\subsection{Real-World Transfer (RQ3)}
\label{subsec:real_world_results}

RQ3 evaluates whether the planning and recovery capabilities established in simulation transfer to a physical mobile manipulator under increasing task difficulty.

\textbf{Real-world setup and baselines.} Experiments use a 3\,m $\times$ 3\,m workspace divided into \textit{Cupboard}, \textit{Dining table}, and \textit{Pantry} regions (Fig.~\ref{fig:MoMaStage_realworld_overview}). These regions provide the map-light prior without using a dense geometric map or scene graph. \pname runs on the Agilex Cobot Magic platform, comprising four Agilex Piper arms, a Tracer mobile base, three Orbbec Dabai manipulation cameras, and an Intel RealSense D435i. The real-robot SSG contains 16 semantic skills (6 pick, 6 place, and 4 navigation) connected by 63 edges, together with 22 action-level recovery skills outside the semantic planning vocabulary. Semantic manipulation skills use ACT trained on 50 spatially generalized demonstrations, while navigation and recovery use predefined rules. We compare GT Sequencing, DeCo*, and \pname under the same backbone and skill vocabulary, together with an end-to-end ACT baseline trained on 50 full-task demonstrations.

\textbf{Difficulty stratification.} We evaluate 100 trials and assign each a pre-execution difficulty score based on horizon length, scene switches, object interactions, receptacle actions, and recovery opportunities. The trials are divided into \emph{easy} (3 scene tasks, 20 trials per scene task, 60 trials in total, average horizon 5), \emph{medium} (3 scene tasks, 10 trials per scene task, 30 trials in total, average horizon 11), and \emph{hard} (1 scene task, 10 trials per scene task, 10 trials in total, average horizon 17). The score is used only for stratification and is not provided to the planner or used to alter the skill library, prompts, or controller policies. Planning validity and task completion are evaluated on the same trials.

\textbf{Real-world results.} On the easy and medium tiers, \pname achieves 100\% pre-execution planning validity and final task-completion rates of 90\% and 80\%, respectively. We focus the comparison on the hard tier, which stresses cross-region navigation, object transfer, and dual-arm manipulation. Figure~\ref{fig:MoMaStage_realrobot_overall} decomposes the results into per-subtask success, trajectory-level execution states for 10 trials, and a representative recovery during subtask 6. An unrecovered failure terminates the episode. End-to-end ACT does not complete the evaluated hard-tier tasks. DeCo* completes the first subtask in 10\% of trials and reaches 0\% survival by subtask 7, while open-loop GT reaches 0\% by subtask 16. \pname maintains 100\% pre-execution planning validity and completes 60\% of the full tasks. Successful trajectories include detected deviations followed by localized recovery. Remaining failures involve objects leaving the reachable workspace or unrecoverable controller states.

These results answer RQ3 affirmatively at the planning and recovery levels: the SSG prevents the measured state-inconsistent skill chains, and event-triggered replanning allows execution to continue without discarding completed subtasks. This behavior is important in long-horizon mobile manipulation because restarting after every failure would repeat the cost of navigation and interactions already completed. The gap between 100\% planning validity and 60\% task completion on the hard tier nevertheless exposes the remaining dependence on physical execution. In successful recoveries, the monitor updates the execution state and repairs only the affected portion of the plan. Conversely, graph-level reasoning cannot compensate for insufficient dexterity, objects outside the reachable workspace, or fundamentally unreachable controller states. The real-world results therefore demonstrate transfer while delineating the recovery boundary of the compact state abstraction.

\section{CONCLUSION}
\label{sec:conclusion}

We present \pname, a map-light framework coupling a frozen VLM to robot execution for state-consistent planning and closed-loop execution in long-horizon indoor mobile manipulation. A hierarchical skill library constrains the planning vocabulary, a Skill-State Graph verifies cumulative scene-region and gripper-occupancy state before execution, and event-triggered monitoring invokes graph-grounded repair when observed outcomes invalidate the remaining plan. In simulation, \pname achieves higher planning validity and intermediate execution survival than the evaluated baselines, while real-robot experiments demonstrate closed-loop recovery on long-horizon tasks. The topology-only interface also reduces latency across backbone comparisons and reasoning-token use on the evaluated reasoning-capable backbone. These results show that explicit embodiment-state transitions provide a practical interface between high-level semantic planning and robot execution, while performance remains bounded by underlying physical skills. Future work will improve the robustness, dexterity, and recovery scope of manipulation and navigation skills and explore automated SSG construction while retaining the lightweight state abstraction and event-driven replanning design.

\bibliographystyle{IEEEtran}
\bibliography{reference}

\end{document}